\documentclass[conference]{IEEEtran}
\IEEEoverridecommandlockouts
\usepackage{cite}
\usepackage{amsmath,amssymb,amsfonts}
\usepackage{algorithm}
\usepackage{algpseudocode}
\usepackage{graphicx}
\usepackage{textcomp}
\usepackage{hyperref} 
\usepackage{xcolor}
\def\BibTeX{{\rm B\kern-.05em{\sc i\kern-.025em b}\kern-.08em
    T\kern-.1667em\lower.7ex\hbox{E}\kern-.125emX}}
\begin{document}
\newcommand{\bmat}[1]{\begin{bmatrix}#1\end{bmatrix}}

\title{Robotic versus Human Teleoperation for Remote Ultrasound}

\author{\IEEEauthorblockN{David Black}
\IEEEauthorblockA{\textit{Electrical and Computer Engineering} \\
\textit{University of British Columbia}\\
Vancouver, Canada \\
dgblack@ece.ubc.ca}
\and
\IEEEauthorblockN{Septimiu Salcudean}
\IEEEauthorblockA{\textit{Electrical and Computer Engineering} \\
\textit{University of British Columbia}\\
Vancouver, Canada \\
tims@ece.ubc.ca}
}


\maketitle

\begin{abstract}
Diagnostic medical ultrasound is widely used, safe, and relatively low cost but requires a high degree of expertise to acquire and interpret the images. Personnel with this expertise are often not available outside of larger cities, leading to difficult, costly travel and long wait times for rural populations. To address this issue, tele-ultrasound techniques are being developed, including robotic teleoperation and recently human teleoperation, in which a novice user is remotely guided in a hand-over-hand manner through mixed reality to perform an ultrasound exam. These methods have not been compared, and their relative strengths are unknown. Human teleoperation may be more practical than robotics for small communities due to its lower cost and complexity, but this is only relevant if the performance is comparable. This paper therefore evaluates the differences between human and robotic teleoperation, examining practical aspects such as setup time and flexibility and experimentally comparing performance metrics such as completion time, position tracking, and force consistency. It is found that human teleoperation does not lead to statistically significant differences in completion time or position accuracy, with mean differences of $1.8\%$ and $0.5\%$, respectively, and provides more consistent force application despite being substantially more practical and accessible.
\end{abstract}


\section{Introduction}
Remote and under-resourced communities have far worse access to healthcare than larger cities \cite{adams2021, adams2022}. Ultrasound has become one of the most prevalent diagnostic imaging modalities due to its relatively low cost, non-invasive nature, and lack of radiation \cite{wu2013}, but many communities have very limited access to qualified sonographers. While ultrasound transducers are accessible, a high degree of expertise is required to acquire images \cite{blehar2015}. Patients must therefore rely on a traveling physician who comes every few weeks, or must themselves travel to a larger city for even routine examinations. This leads to long wait times, high cost, work and family interruptions, and an inability to manage emergency cases effectively \cite{adams2023}.

Tele-ultrasound has emerged as a valuable tool for some cases, diverting travel and bringing expert knowledge to the communities virtually. However, current methods have significant drawbacks. Most point-of-care ultrasound (POCUS) systems rely on a remote expert seeing the ultrasound image live and giving verbal feedback or instructions to the operator~\cite{hermann2022, nelson2016}. This is very convenient but inefficient and imprecise for all but an experienced operator who is already competent in ultrasound. Video tele-guidance systems are thus not widely used for truly performing remote ultrasound exams. 

Conversely, after its first application in 1999 \cite{salcudean1999, salcudean2000}, robotic teleultrasound has recently experienced a resurgence in commercial interest, with companies and start-ups including AdEchoTech \cite{adams2022a}, Dopl \cite{dopl}, Life Science Robotics \cite{lsr}, MGI \cite{mgi}, and Cobionix \cite{cobionix} each pursuing similar systems. Telerobotics has the advantage of giving the expert full control, with good precision and efficiency \cite{mathiassen2016,wang2019,wang2021echo}. Much recent research has also explored AI for autonomous robotic ultrasound \cite{salcudean2022, jiang2023}. However, telerobotics equipment is incongruously expensive, large, inflexible, and complex compared to POCUS devices. It is also more difficult for patients to accept than a human operator and often requires complex calibration and registration before use. These factors limit the accessibility and practicality of robotic teleultrasound systems for small communities.

Human teleoperation has been proposed as a potential compromise between the efficiency and precision of robotics with the simplicity and ease-of-use of verbal teleguidance \cite{black2023hci,black2024cag}. In human teleoperation, the robot of a telerobotic system is replaced by an inexperienced follower person who tracks the motion of a virtual ultrasound probe shown in a mixed reality (MR) headset. As in many conventional teleoperation systems, the expert inputs their desired motion through a haptic device, receives force feedback, and sees the ultrasound image and a video stream in real time. The virtual probe in the follower's headset carries out the expert's motion and is tracked by the follower, who thus carries out the scan.

It has been shown that human teleoperation controller design can be treated control theoretically like a telerobotic system \cite{black2024bilat,black2025linear}, and that the follower person can track the virtual probe with an accuracy of approximately 3 mm, $6^\circ$, $0.3$ N, in position, orientation, and force tracking, respectively, and a lag of 350~ms \cite{black2023ijcars}. This led to much improved performance compared to verbal teleguidance in preliminary tests \cite{black2023hci}. Moreover, the system has shown practical utility in tests over a distance of 750km in a remote, Indigenous community \cite{yeung2024}. However, it is unclear how such a system compares to robotic teleoperation in terms of task performance. 
\begin{figure*}[t]
    \centering
    \includegraphics[width=0.7\linewidth]{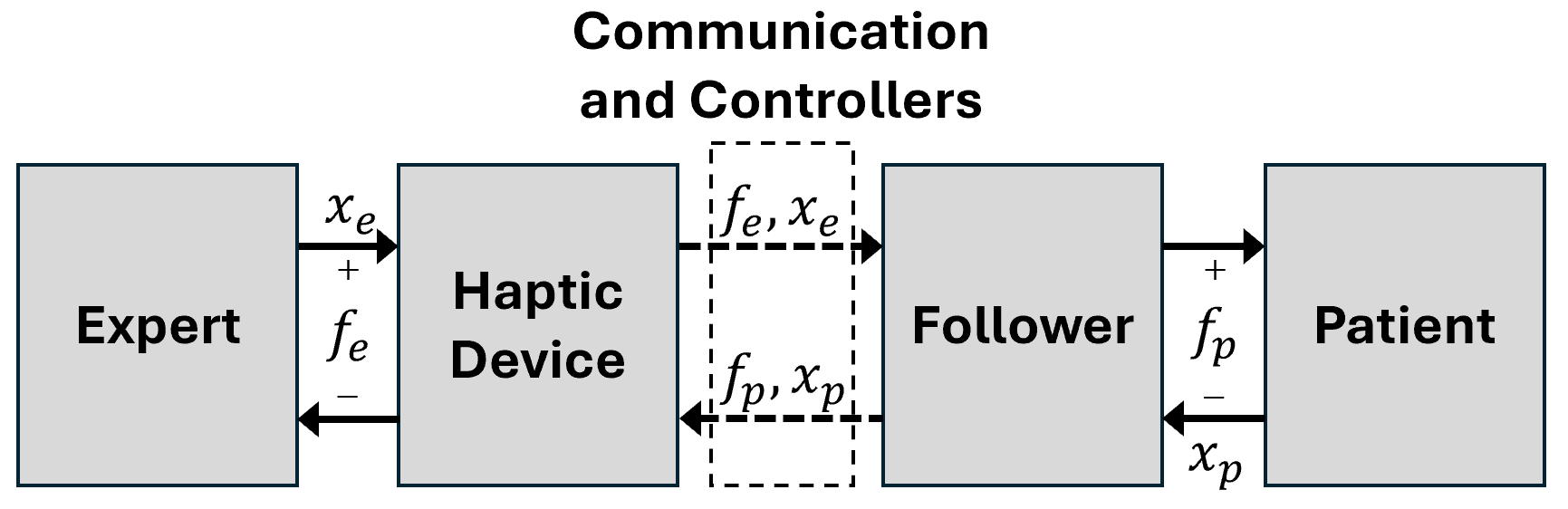}
    \caption{General diagram of teleoperation for ultrasound. The follower can be a robot or a person with a mixed reality headset. The expert interacts with a haptic device to input their desired motion and receive force feedback, while the follower interacts with the patient. Force and position are interchanged between the two pairs.}
    \label{fig:overall}
\end{figure*}

Raw performance values such as tracking accuracy and speed can be compared from prior studies. However, we found in human testing that absolute accuracy was less relevant for human-teleoperated ultrasound as the expert made relative motions based on the ultrasound image alone \cite{yeung2024}. The same is likely true in robotic ultrasound. Furthermore, a robotic system has to be safe for the patient and operator and stable in intermittent contact with the patient surface. Impedance controllers are best suited for such scenarios due to their compliant behavior \cite{dyck2024}. As a result, however, the robot's position is necessarily offset from the expert's input whenever in contact with the patient. Therefore, comparing pose tracking accuracy is relatively meaningless. A more important criterion is the ability to track a feature such as a blood vessel in ultrasound image space.

Similarly, the tracking lag of the robot or follower person is important as it determines how responsive and intuitive the system feels for the expert. Human teleoperation is limited by the reaction time of the follower person, who must see, process, and respond to the changing virtual probe \cite{black2023ijcars, black2024tmrb}. Conversely, while robot control can be very fast, human-robot interaction considerations necessitate a slower, more gentle behavior. The velocity and acceleration must be substantially limited to avoid impacting the patient or operator with the robot. As a result, a certain tracking delay is present in the robot as well. The amount of delay, however, depends on how much the velocity is limited, which makes it a question of design and judgment of acceptable safety margins rather than a performance limitation. Thus, tracking lag is difficult to compare.

The ability of the expert to control the applied force is perhaps more relevant. An impedance-controlled robot will push harder if the expert pushes harder, whereas a follower person must be induced to apply more force through visual cues such as moving the virtual probe deeper into the patient \cite{black2024bilat}. The latter may be less reliable. As shown in recent robotic ultrasound research, however, maintaining a consistent force on a non-uniform surface while scanning with an ultrasound probe is still not easily achieved \cite{dyck2022}, while it is relatively easy for a person. Thus, the relative force tracking ability is unknown.

Furthermore, as alluded to above, raw performance metrics such as tracking accuracy do not necessarily translate directly to good performance on an ultrasound scanning task, and practical considerations such as setup time, cost, and complexity are equally important for clinical translation. 

This paper therefore aims to compare these factors, as well as force tracking ability and the tracking accuracy in ultrasound image space, between human and robotic teleoperation. To this end, we first introduce the human teleoperation system and an analogous robotic teleoperation system before describing the comparison experiments. The two teleoperation methods are additionally compared to direct scanning as a baseline. Using an ultrasound phantom for repeatability, the following questions are addressed quantitatively and the results are compared for the two systems, as well as direct scanning:

\begin{enumerate}
    \item \textit{Tracking ability}: How accurately can the expert follow a desired trajectory in ultrasound image space?
    \item \textit{Force control}: How consistent can the expert keep the applied force on the phantom?
    \item \textit{Task completion}: Can the expert successfully complete five set tasks, and what quality do they achieve?
    \item \textit{Speed}: What is the completion time for the five tasks?
    \item \textit{Complexity}: How complex and time-consuming is the system to set up?
\end{enumerate}

\section{Methods}\label{sec:meth}

\begin{figure*}[t]
    \centering
    \includegraphics[width=\linewidth]{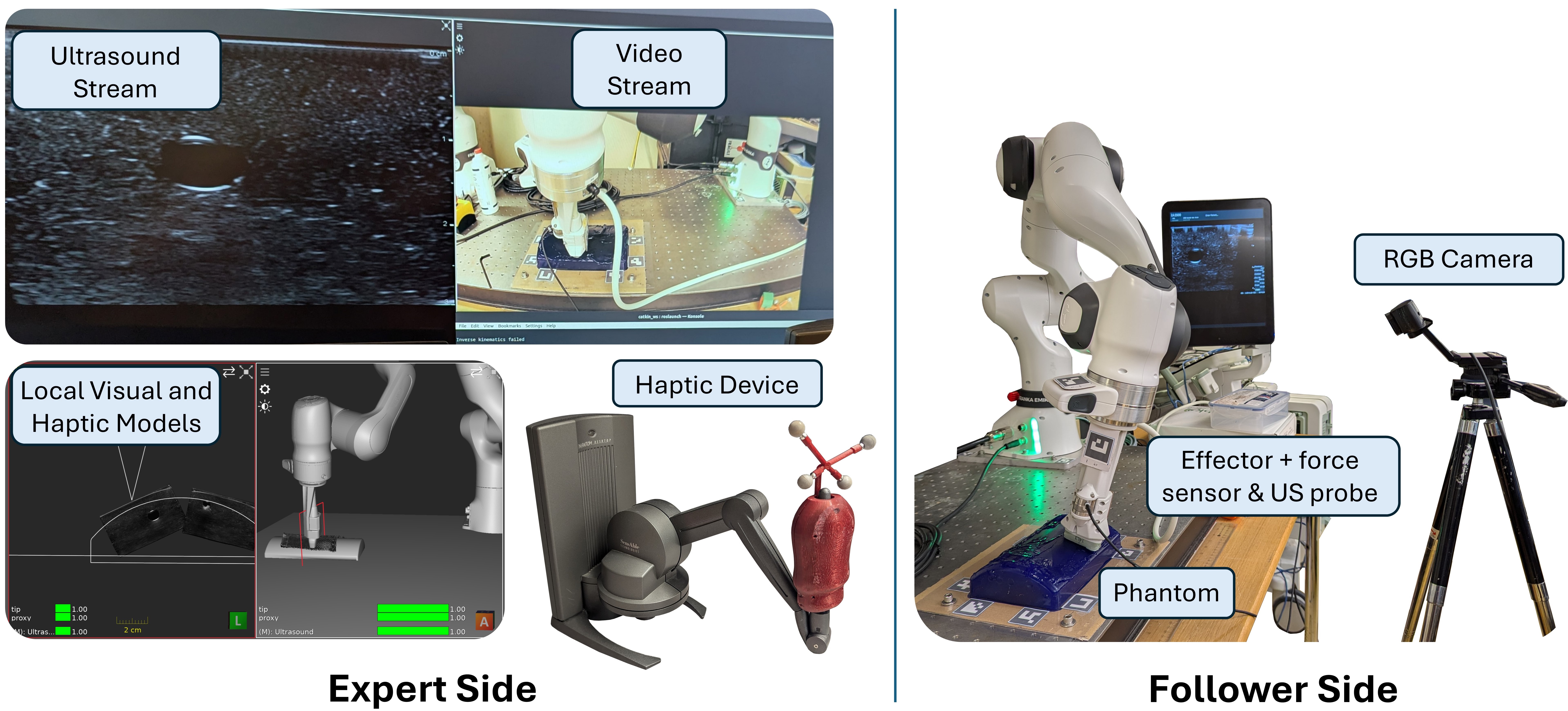}
    \caption{Overview diagram of the robotic teleoperation system, showing the expert (left) and follower (right) sides, with haptic device and visualization for the expert, and a Franka Panda robot with BK ultrasound probe and tissue phantom on the follower side. }
    \label{fig:roboDiagram}
\end{figure*}

This section describes the robotic and human teleoperation systems and the experiments carried out to compare their effectiveness for teleultrasound. Both systems were designed to be as similar as possible, fitting into the same general teleoperation architecture shown in Fig. \ref{fig:overall} with the follower person as a drop-in replacement for the follower robot.

\subsection{Telerobotic Ultrasound System}
The robotic teleoperation setup was similar to the one previously described in \cite{black2025iros}, where more detail is provided about the controller. It is illustrated in Fig. \ref{fig:roboDiagram}. The teleoperation consists of the expert side interface and the follower robot, which communicate using Web Real Time Communication (WebRTC). This is a fast peer-to-peer framework built on stream control transmission protocol (SCTP) for data and secure real-time transmission protocol (SRTP) for video and audio. During the tests described below, the round-trip delay was consistent at approximately $5\pm4$ ms on the local WiFi network.

The expert side consisted of a desktop application implemented in C++ and the ImFusion Suite (ImFusion GmbH, Munich, Germany), and a Touch X haptic device (3D Systems, Rock Hill, SC) with a dummy ultrasound probe as an end effector. The haptic device was used to obtain the expert's input motion, and force feedback was provided through a virtual fixture. The virtual fixture was implemented using a triangle mesh of the ultrasound phantom (see Section \ref{ssec:exp}), made in SolidWorks and placed in the ImFusion Suite at the correct transform relative to the follower robot. In practice, this can be replaced by a mesh of the patient from an RGB-D camera or similar. Forces were computed using the proxy point method \cite{ruspini1997} and a virtual spring-damper preventing the end effector from entering the volume defined by the phantom mesh. The spring and damping constants were manually tuned to approximate the mechanical impedance of the ultrasound phantom. In addition to force feedback, the expert interface showed the live ultrasound and video streams from the follower side.

These streams came from the ultrasound probe and an RGB camera mounted facing the robot, respectively. The ultrasound probe was a BK Medical 14L3 linear transducer mounted through a custom PLA 3D printed end effector on the flange of a Panda robot (Franka Robotics, Munich, Germany). The robot has seven degrees of freedom and ran an impedance controller for compliant interaction with the phantom, using the PREEMPT\_RT real-time kernel on Ubuntu, together with ROS and libfranka. Desired poses from the haptic device were sent to the follower host PC and transformed using a predetermined and fixed expert to follower transform before computing the inverse kinematics using the trac\_ik library \cite{beeson2015}. Since the manipulator is redundant, the inverse kinematics was run until it found multiple solutions, using Newton-Raphson iteration and sequential quadratic programming, and the solution resulting in a joint configuration closest to the current configuration was selected. This was then passed to a faster thread, where Ruckig \cite{berscheid2021} was used to interpolate between successive pose commands in a jerk and acceleration-limited manner. Communication between threads was achieved using fast, lock-less, thread-safe queues\footnote{\href{https://github.com/cameron314/readerwriterqueue}{https://github.com/cameron314/readerwriterqueue}}. The commanded joint positions were maintained using a computed torque PD controller with dynamics feed-forward including friction, Coriolis, and gravity compensation based on libfranka. The proportional and derivative terms are equivalent to stiffness and damping and were tuned for compliant yet responsive performance.

\subsection{Human Teleoperation System}
The human teleoperation system has been described in previous papers \cite{black2023vr}. In brief, it consists of two sides - the expert and follower - connected by a communication system also built on WebRTC. Both the expert side and communication were designed to be identical to the telerobotic system, to isolate solely the effects of the human versus robotic follower. Thus, the expert manipulated a Touch X haptic device with a dummy ultrasound probe end effector and received force feedback from a virtual fixture derived from the mesh of the phantom, with stiffness and damping tuned to approximately match the real phantom. The live ultrasound image stream and a video from the follower's point of view camera on the mixed reality headset were also shown. 

\begin{figure}[h]
    \centering
    \includegraphics[width=0.7\linewidth]{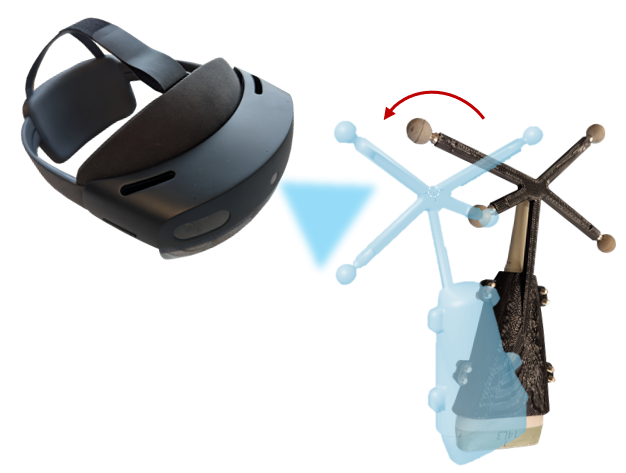}
    \caption{Human teleoperation system, in which the follower aligns the real ultrasound probe with a virtual one (blue in the image) in a mixed reality headset. The expert side interface and communication system are the same as in Fig. \ref{fig:roboDiagram} for the robotic teleoperation.}
    \label{fig:humanTeleop}
\end{figure}

On the follower side, the novice follower person wore the mixed reality headset, a HoloLens 2 (Microsoft, Redmond, WA), which rendered a 3D virtual ultrasound probe. The virtual probe's pose was controlled in real time by the expert's haptic device, and the follower's only task was to align the real probe to the virtual one, as shown in Fig. \ref{fig:humanTeleop}. To avoid any bias from some followers being more or less familiar with ultrasound imaging, the followers were blinded to the ultrasound image and acted simply as position-controlled robots. The implications of this are discussed in Section \ref{sec:discussion}. The HoloLens application was developed in Unity and ArUco markers \cite{garrido2014} on the phantom base plate were used to align the HoloLens scene to the expert's coordinate system.

Since the virtual fixture's mechanical impedance approximately matched that of the phantom, force control was achieved simply by the follower tracking the virtual probe. A downward force on the virtual fixture by the expert results in the haptic device handle moving down, so the virtual probe moves down by the same amount, and in order to match its position, the follower must apply the same force to the phantom. For simplicity, no additional feedback in position or force was given to the follower, as described in other work \cite{black2024bilat}, though these may improve performance.

\subsection{Experiments}\label{ssec:exp}
To compare human and robotic teleoperation for ultrasound, volunteer subjects performed scans with both systems introduced above, as well as directly with the probe in their hand. For repeatability, a Blue Phantom branched 2-vessel training block (CAE Healthcare, Inc.) was used as the subject. The phantom was fixed to a board with ArUco markers \cite{garrido2014} and attached to the same optical prototyping board as the Franka robot. The 3D mesh model was measured prior to testing and saved with the correct transform relative to the follower robot. This was loaded into the expert-side application for each test to save time and ensure repeatability. In a practical situation, the point cloud can update in real-time or be re-scanned before every exam so it is patient specific.

\begin{figure}[h]
    \centering
    \includegraphics[width=\linewidth]{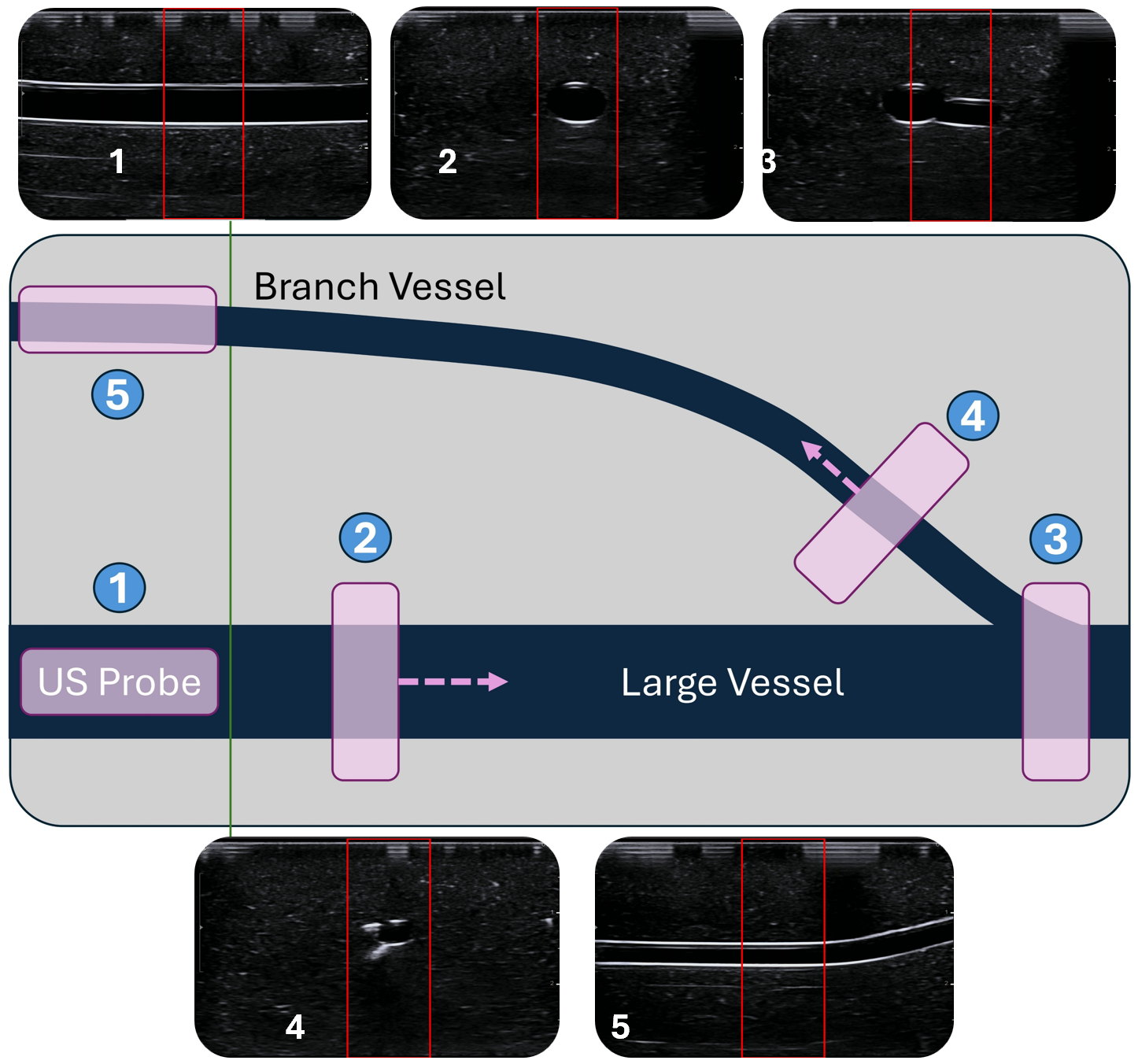}
    \caption{Diagram of the ultrasound scanning task on the branching vessel phantom. The typical ultrasound images from steps 1-5 are shown, and the steps are described in the text. The red lines in the ultrasound images show the subjects where to keep the vessels centered during the sweeps.}
    \label{fig:testPlan}
\end{figure}

The order of the human teleoperation, robotic teleoperation, and direct scanning was randomized to avoid learning effects, and the volunteers were given 5 minutes prior to testing to become familiar with each interface. The ultrasound scanning task consisted of the following steps, shown also in Fig. \ref{fig:testPlan}.
\begin{enumerate}
    \item Find a clean longitudinal view of the left end of the large vessel and hold it for 5 seconds.
    \item With a transverse view, sweep along the length of the large vessel, keeping it centered in the image.
    \item Center the bifurcation of the two vessels and hold it for 5 seconds.
    \item Sweep along the thin vessel, keeping it centered.
    \item Find a longitudinal view of the left end of the thin vessel and hold it for 5 seconds.
\end{enumerate}

A line drawn on the phantom's board indicates the minimum extent of the sweeps, and the point at which the longitudinal views must be captured. This is indicated in green in Fig.~\ref{fig:testPlan}. During the tests, the screen showing the ultrasound image and the RGB stream was recorded, and a box was rendered on the ultrasound image to indicate its center. The users were instructed to keep the vessels centered laterally within the box during the sweeps, to maintain a constant force, and to finish as quickly as possible.

After the tests, the recorded images were evaluated. The vessel was first segmented using simple thresholding, opening, and closing operations in MATLAB, and the deviation of its centroid from the image center was determined for every frame of the sweep. In addition, the width, $w$, and height, $h$ of the vessel were measured and used to determine the eccentricity:
$$e=\sqrt{1-\dfrac{h^2}{w^2}}$$
Finally, the completion time was recorded for each sub-task and the procedure as a whole.

The tests involving participants in this study were performed in line with the principles of the Declaration of Helsinki and approval of the University of British Columbia Behavioural Research Ethics Board (Approval No: H22-01195). Informed consent was obtained from all participants.

\section{Results}\label{sec:res}
Fifteen volunteers (5 female; mean age 27) acted as experts for human and robotic teleoperation, and alternated being the follower for each other in the human teleoperation. Each user also performed the ultrasound scan directly. The results are shown in the following tables, with p-values from the two-sample Komogorov-Smirnov test. 

\begin{table}[h]
\centering
\caption{Completion times, in seconds, comparing human and robotic teleoperation, and direct scanning.}
\begin{tabular}{|c|c|c|c|}\hline
 & Vessel Finding & Sweeping & Total\\\hline
Human & $68\pm25$ & $162\pm28$ & $229\pm26$\\\hline
Robotic & $73\pm112$ & $152\pm33$ & $225\pm133$\\\hline
Direct & $46\pm74$ & $150\pm65$ & $196\pm64$\\\hline
p (Human-Robotic) & 0.87 & 0.55 & 0.95\\\hline
p (Human-Direct) & 0.15 & 0.40 & 0.02\\\hline
\end{tabular}
\label{tab:humanSpeed}
\end{table}
Table \ref{tab:humanSpeed} shows the completion times, with human teleoperation outperforming telerobotics in finding the longitudinal vessel planes, but going slower in the sweeping tasks. Overall, the completion time was similar, differing by $1.8\%$ and showing no statistical significance. The discrepancy between the two methods on the two different tasks may be because the human follower moved more carefully and tended to lag the expert motion slightly, which made it easier for the expert to find small features without overshooting, but took longer for sweeps. Direct ultrasound was significantly faster in total, but had $p>0.05$ in the individual stages. With direct ultrasound, there was no delay between the operator seeing the longitudinal vessel view and being able to stop the probe, making the vessel finding task faster in particular.

\begin{table}[h]
\centering
\caption{Image-space tracking accuracy (RMSE in pixels) between human and robotic teleoperation and direct scanning.}
\begin{tabular}{|c|c|c|c|}\hline
Method & Error & p-Value vs. Robotic & p-Value vs. Human \\\hline
Human & 185 & 0.03 & - \\\hline
Robotic & 186 & - & 0.03\\\hline
Direct & 183 & $<0.01$ & 0.05\\\hline
\end{tabular}
\label{tab:trackAccuracy}
\end{table}
The ultrasound-image-space tracking accuracy is shown in Table \ref{tab:trackAccuracy}. The expert's ability to keep the vessels centered during sweeps was almost exactly the same in the human and robotic teleoperation, and only slightly improved in direct ultrasound. All differences were statistically significant, though very minor, and the p-value for human teleoperation versus direct ultrasound was 0.05.

\begin{figure}[h]
    \centering
    \includegraphics[width=\linewidth]{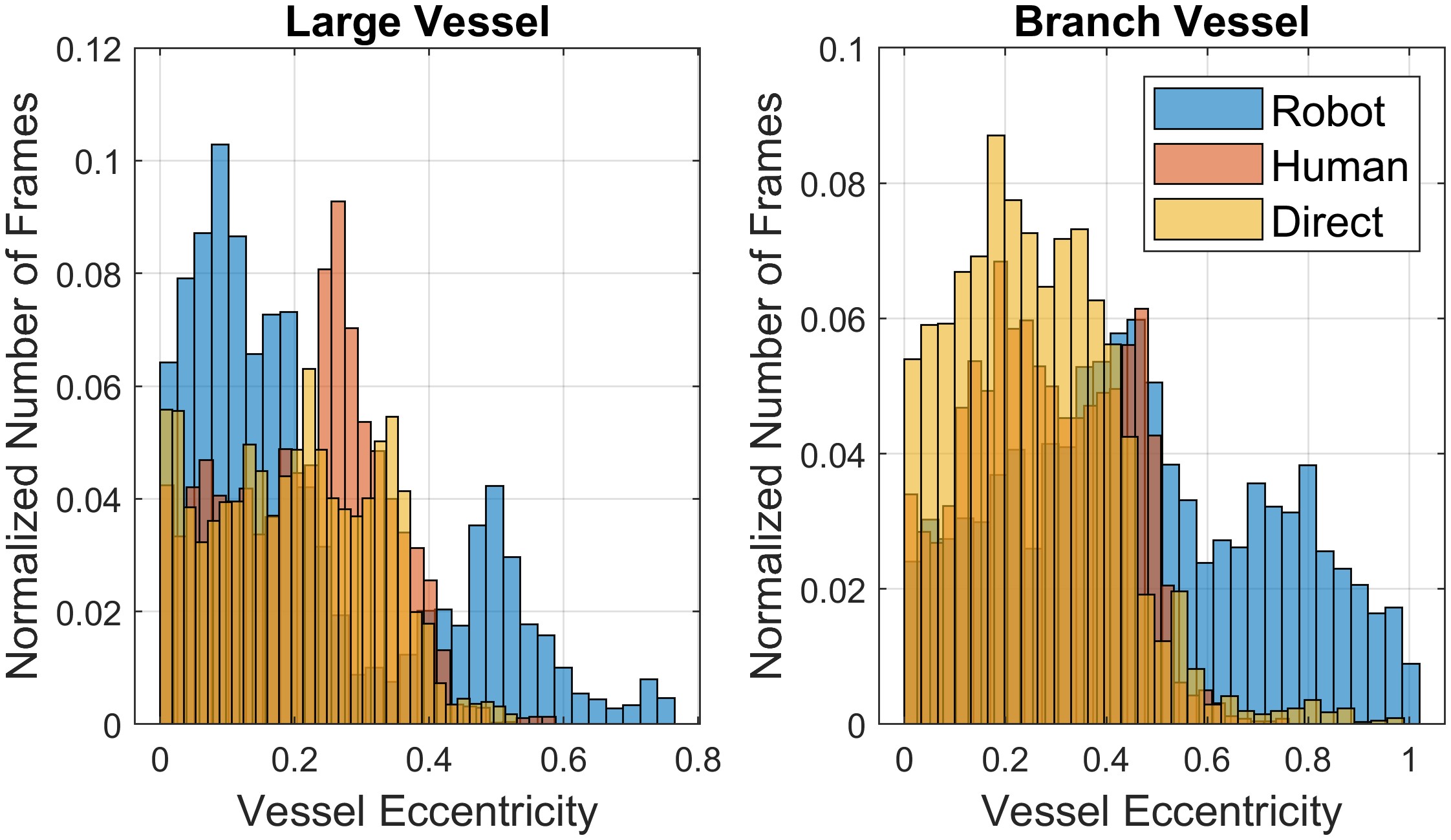}
    \caption{Vessel eccentricity distribution, showing the magnitude and consistency of the applied force on the large and branch vessel sweeps. Robotic teleoperation shows a larger tail towards an eccentricity of 1, indicating higher and less consistent forces.}
    \label{fig:eccentricity}
\end{figure}

Conversely, the applied force is shown in Table \ref{tab:force} and Fig.~\ref{fig:eccentricity}, in the form of vessel eccentricity. A higher eccentricity indicates more vessel compression, and thus a higher applied force. The human follower maintained a significantly more consistent force on the phantom than the robotic arm and had lower overall force amplitude, both of which are important for patient comfort. Direct ultrasound in turn performed slightly better than human teleoperation.

\begin{table}[h]
\centering
\caption{Vessel eccentricity, showing the magnitude and consistency of the applied force on the large and branch vessel sweeps. All differences are significant ($p<0.001$)}
\begin{tabular}{|c|c|c|c|}\hline
 & Human & Robotic & Direct\\\hline
Large & $0.21\pm0.12$ & $0.22\pm0.18$ & $0.20\pm0.12$\\\hline
Branch & $0.28\pm0.15$ & $0.47\pm0.26$ & $0.26\pm0.16$\\\hline
\end{tabular}
\label{tab:force}
\end{table}

\section{Discussion and Conclusion}\label{sec:discussion}
The results show that human teleoperation does not sacrifice performance compared to robotic teleultrasound. Although telerobotics was slightly faster on average, the difference was not statistically significant and robotics had much greater variance in completion time. In terms of pose tracking accuracy, i.e. the ability to follow a desired trajectory in ultrasound image space, both methods performed almost identically. Human teleoperation also led to more consistent application of force, which is more comfortable for the patient. This is dependent on the robot controller, but in general applying a given normal force during intermittent contact while moving laterally on a curved surface is difficult for robots \cite{dyck2022} but trivial for people. The human arm naturally has much lower impedance than the Franka robot, so compliant behavior is easier to achieve.

Direct ultrasound outperformed both teleoperation methods. This is expected, though the trade-off is of course that the two teleoperation methods enable tele-ultrasound at a distance, whereas direct ultrasound does not solve the problem of ultrasound access in remote areas. Moreover, the differences in performance were relatively small. However, larger differences would likely be seen in practical tasks such as positioning the patient initially or in pressing dynamically on a real patient to move organs or abdominal gas, for example, though this remains to be shown in future work.

In addition to achieving very similar performance, the two teleoperation techniques provided a remarkably similar expert experience while performing the tests. Both function with an identical user interface and have similar tracking behavior, making them almost indistinguishable from the expert's perspective. 

The human teleoperation performance could potentially be improved further by showing the follower the ultrasound image as well. In this case, if the human follower knows approximately what view is desired, they are be able to slow down or stop at the plane even if the expert overshoots slightly due to time delay, whereas the robot simply overshoots as well. This amounts to a shared control between the intelligent follower and the expert that is relatively difficult to achieve with a robot. In this way, the primary differences between the two methods are that the human follower may sometimes not follow perfectly if distracted, but can act intelligently and follow verbal commands as well.

There are also other important practical factors such as cost, portability, and setup time to consider. When setting up the robot, it was necessary to switch the robot on, start a ROS node, and align the orientation of the haptic device with the robot's. This took on average 1:53 minutes. Conversely, in human teleoperation it was necessary to turn on the HoloLens~2, start the application, hold the US probe on the phantom, and press a ``Start" button. This took 0:34 on average. Moreover, this does not include moving the robot into position, fixing it securely in place, and performing the hand-eye and robot-US calibrations, all of which are very time-consuming, must be completed with precision, and are not required for human teleoperation. 

Regarding cost, a Franka Panda robot, as used in our system, costs approximately $\$37000$ Canadian Dollars, while the HoloLens~2 was purchased for $\$5000$ CAD and substantially lower cost MR glasses are also available. For example, the XReal Air 2 Ultra costs $\$1000$ CAD and can support human teleoperation. In both human and robotic teleoperation, a local operator is needed to set up the robot or act as follower, respectively. Furthermore, having a human follower is safer than a robot as the human hand will never perform a sudden, dangerous motion, apply excessive force, or become unstable. Near singularities or joint limits, the behavior of a robot must be considered very carefully. People can also react better to unforeseen circumstances than robots, and the system is small and mobile, making it easy to move between rooms, buildings, or even to emergency sites for scans.

Further study with more subjects, more expert sonographers, and a remote setting could strengthen the results. In addition, future tests should use commercially available robotic systems such as the Melody robot from AdEchoTech, and compare clinical measures of image quality and diagnostic value from real patient scans.

This paper has shown that for an ultrasound exam, human and robotic teleoperation lead to very similar performance and results, despite the much lower cost and higher practicality, simplicity, and ease-of-use of human teleoperation.

\bibliographystyle{ieeetr}
\bibliography{refs}
\end{document}